\documentclass[sigconf]{acmart}
\AtBeginDocument{%
  }

\usepackage{algorithm}
\usepackage{svg}

\copyrightyear{2026}
\acmYear{2026}
\setcopyright{cc}
\setcctype{by}
\acmConference[RecSys '26]{20th ACM Conference on Recommender Systems}{September 27-October 02, 2026}{Minneapolis, MN, USA}
\acmBooktitle{20th ACM Conference on Recommender Systems (RecSys '26), September 27-October 02, 2026, Minneapolis, MN, USA}
\acmDOI{10.1145/3773078.3831919}
\acmISBN{979-8-4007-2284-4/2026/09}

\begin{document}

\title[Self-Evolving Recommendation System]{Self-Evolving Recommendation System: End-To-End Autonomous Model Optimization With LLM Agents}
\author{Haochen Wang}
\authornote{Equal contribution to the work.}
\affiliation{
  \institution{Google Inc}
  \city{Mountain View}
  \state{California}
  \country{USA}
}
\email{haochenww@google.com}

\author{Yi Wu}
\authornotemark[1]
\affiliation{
  \institution{Google Inc}
  \city{Mountain View}
  \state{California}
  \country{USA}
}
\email{wuyish@google.com}

\author{Daryl Chang}
\authornotemark[1]
\affiliation{
  \institution{Google Inc}
  \city{Mountain View}
  \state{California}
  \country{USA}
}
\email{dlchang@google.com}

\author{Li Wei}
\affiliation{
  \institution{Google Inc}
  \city{Mountain View}
  \state{California}
  \country{USA}
}
\email{liwei@google.com}

\author{Lukasz Heldt}
\affiliation{
  \institution{Google Inc}
  \city{Mountain View}
  \state{California}
  \country{USA}
}
\email{heldt@google.com}

\begin{abstract}
Optimizing large-scale machine learning systems, such as recommendation models for global video platforms, requires navigating a massive hyperparameter search space and, more critically, designing sophisticated optimizers, architectures, and reward functions to capture nuanced user behaviors. Achieving substantial improvements in these areas is a non-trivial task, traditionally relying on extensive manual iterations to test new hypotheses. We propose a self-evolving system that leverages Large Language Models (LLMs), specifically those from Google's Gemini family, to autonomously generate, train, and deploy high-performing, complex model changes within an end-to-end automated workflow. The self-evolving system consists of an Offline Agent (Fast Loop) that performs high-throughput hypothesis generation to optimize for proxy metrics, and an Online Agent (Slow Loop) that validates candidates against delayed north star business metrics in live production. Our agents act as specialized Machine Learning Engineers (MLEs): they exhibit deep reasoning capabilities, discovering novel improvements in optimization algorithms and model architecture, and formulating innovative reward functions that target long-term user engagement. The effectiveness of this approach is demonstrated through several successful production launches at YouTube, confirming that autonomous, LLM-driven evolution can surpass traditional engineering workflows in both development velocity and model performance.
\end{abstract}



\maketitle

\section{Introduction}
Global video platforms like YouTube serve billions of users by curating personalized feeds from vast corpora of content. At the core of delivering relevant experiences is the recommendation system, an ensemble of algorithms and models designed to help users discover content they love. Increasingly, modern recommendation systems are being formulated as Reinforcement Learning (RL) problems \cite{reinforce_paper, deeprl_paper}, where the system acts as an agent interacting with a user environment to maximize cumulative utility over time. As surveyed in \cite{rl_survey}, this paradigm shifts the focus from simple Click-Through Rate (CTR) prediction to optimizing long-term user satisfaction, requiring models to balance immediate gratification with delayed rewards like retention and diverse content exploration.

However, a critical bottleneck in this paradigm is the alignment gap between training proxies and long-term user satisfaction. While models are trained on differentiable loss functions, the actual goal is user satisfaction, which is non-differentiable, delayed, sparse, and often semantically complex. Recent approaches like the Learned Ranking Function \cite{lrf_paper} attempt to bridge this gap by parameterizing the reward function itself, allowing the system to learn the optimal trade-off between conflicting objectives. Similarly, work on diversifying by intent \cite{intent_paper} highlights that modern reward functions must now encode nuanced psychological concepts -- such as user intent and exploration -- rather than simple binary labels.

Optimizing these increasingly semantic and structural components exceeds the capabilities of traditional Automated Machine Learning (AutoML) \cite{automlsurvey_paper}. Standard AutoML methods \cite{bayesian_paper, nas_paper} excel at tuning numerical hyperparameters within fixed search spaces. Yet, they lack the reasoning capabilities to invent new reward logic or architect novel interaction layers from scratch. They cannot interpret past experiment results, hypothesize that a specific user slice is under-served, and write the logic to fix it.

This limitation has catalyzed a shift in the broader machine learning community from "automated tuning" to "autonomous scientific discovery." Recent works such as \cite{scientist_paper, alphaevolve_paper} introduce the concept of AI agents capable of orchestrating the full scientific lifecycle: generating hypotheses, writing code, and refining theories based on empirical results. Unlike rigid AutoML pipelines, these agents utilize Large Language Models (LLMs) to reason over unstructured context. This capability offers a potential solution to the limitation of traditional methods, promising a transition from mere parameter tuning and selection, to automated discovery of complex, novel model changes.

Despite these parallel advancements, a significant gap remains at their intersection. Optimizing industrial-scale recommendation models proves exceptionally difficult and remains a manual, human-intensive endeavor. To bridge the gap, we introduce a \textbf{Self-Evolving Recommendation System} deployed at YouTube. By integrating recent advancements in LLMs with a production recommendation system, we demonstrate the effectiveness of building a rigorous framework where agents act as expert Machine Learning Engineers (MLEs) to solve global-scale open-ended recommendation modeling problems. These agents do not just tune parameters; they read production code, propose structural changes to neural topologies, and formulate complex logic for reward functions within an end-to-end autonomous pipeline. While our primary deployment utilizes Gemini 2.5 Pro \cite{gemini25_paper}, our ablation studies (Section~\ref{sec:gemini_benchmark}) also evaluate a lightweight Gemini variant to quantify the relationship between model reasoning power and discovery performance.

Our contributions are summarized as follows:
\begin{itemize}
    \item \textbf{Autonomous MLE Framework for Industrial-Scale Systems} We introduce a hierarchical agentic system where specialized LLM agents act as expert MLEs to evolve recommendation models. We detail the system design that enables agents to safely manage the full lifecycle of industrial model development -- from hypothesis generation and code implementation to A/B testing.
    \item \textbf{Semantic Discovery} We demonstrate that LLM-based agents can move beyond simple parameter tuning to discover novel architectural components and multi-objective reward functions that align better with long-term user satisfaction, areas previously accessible only to human experts in the recommendation domain.
    \item \textbf{Acceleration of Experimental Velocity} We confirm the success of an autonomous LLM-based evolutionary recommendation system in accelerating the velocity of experimentation and delivery of notable metric gains. With extensive offline and online experiments and production deployments, we validate that agentic systems can surpass hand-tuned baselines and effectively evolve the state-of-the-art in recommendation systems at YouTube.
\end{itemize}

\section{Related Work}
Our work sits at the intersection of automated machine learning, autonomous agents, and RL for recommendation. We distinguish our contributions by contrasting them with existing paradigms in these areas.

\subsection{Automated Model Optimization}
Industry standard practices for automated model optimization in recommendation rely heavily on Hyperparameter Optimization (HPO) \cite{hpo_paper}. Frameworks like Google Vizier \cite{vizier_paper} and Auto-Sklearn \cite{autosklearn_paper} utilize iterative search methods (e.g., Bayesian optimization \cite{bayesian_paper}, Gaussian processes \cite{gaussian_paper}) to tune continuous hyperparameters. However, these methods are confined to defined parameter ranges and lack the semantic depth to interpret why a configuration succeeds or fails. 

Similarly, techniques like optimizer search \cite{lion_paper} and Neural Architecture Search (NAS) \cite{nas_paper} (e.g., DARTS \cite{dart_paper}, evolutionary search \cite{evolutionarc_paper}) apply these principles to make structural improvements. Yet, they remain fundamentally restricted by the constraints of their search spaces, as they can only select or remix from a predefined menu of operations. Consequently, they lack the creative capacity to invent novel modules, refactor code to fix bottlenecks, or introduce complex logic that was not explicitly programmed into the search space.

\subsection{LLMs and Autonomous Agents for Scientific Discovery}
The emergence of LLMs has enabled a shift from selection to generation. Optimization by PROmpting (OPRO) \cite{opro_paper} demonstrates that LLMs can serve as evolutionary operators, iteratively refining solutions based on natural language descriptions. This capability is amplified by models such as Gemini 2.5 \cite{gemini25_paper} with advanced reasoning capabilities, building on the foundations of Chain-of-Thought \cite{cot_paper} and long-context understanding and thinking. 

Beyond optimization, the field of AI agents has exploded with frameworks like ReAct (Reasoning + Acting) \cite{react_paper} and Toolformer \cite{toolformer_paper}, which demonstrate that LLMs can solve complex tasks by interleaving reasoning traces with external tool execution. Building on this, recent "scientist" agents have attempted to automate open-ended workflows. Voyager \cite{voyager_paper} and MetaGPT \cite{metagpt_paper} introduce agents that write executable code to solve open-ended problems, maintaining a persistent repository of reusable functions to accelerate future tasks. AlphaEvolve \cite{alphaevolve_paper}, The AI Scientist \cite{scientist_paper}, and MLE-STAR \cite{mlestar_paper} extend this to algorithmic discovery, where agents perform direct edits on source code to improve performance on academic benchmarks. Our work adapts this "scientist" paradigm to the industrial recommendation system setting. Unlike prior works that optimize for static datasets and academic benchmarks (e.g., ImageNet, Kaggle), our framework addresses the unique challenges of a live production ecosystem: noisy feedback loops, strict safety guardrails, complex user-system interactions, and the need for rigorous A/B testing protocols.

\subsection{Reward Engineering for Reinforcement Learning}

In RL, designing the reward function is often the hardest part of the problem. Eureka \cite{eureka_paper} pioneers the use of LLMs for evolutionary reward design in robotics, while LEARN-Opt \cite{learnopt_paper} optimizes rewards without predefined metrics by using LLMs as analysts to evaluate candidates. However, critical distinctions remain: the availability of a clear oracle for success, and the latency of the feedback loop. While prior approaches are evaluated on robotics or simulations that offer immediate feedback, recommendation systems lack a clear oracle for user satisfaction. The true objective is a latent variable observable only through the delayed, noisy, and sparse real-world interactions on the order of $\Theta(days)$ or $\Theta(weeks)$. Consequently, when designing a reward, we cannot simply optimize for a simulation score; we must instead reason about alignment with offline proxy signals analyzed over petabytes of interaction logs, and ultimately validate our designs through real-world deployment. 

\section{Problem Formulation}
We begin by introducing the components of an RL-based recommendation model, specifically for the task of ranking videos to maximize the total expected long-term user satisfaction. We formulate this task as a bi-level optimization problem. 

Our ultimate goal is to maximize a non-differentiable, long-term user satisfaction metric, which is observable only through online interaction. However, directly optimizing online metrics is intractable because feedback is sparse, delayed, and noisy. Therefore, the problem is bi-level: in the lower level, a ranking model is trained to optimize an engineered proxy objective (the cumulative reward). In the upper level, we find the optimal system configuration (e.g., optimizer, architecture, reward) such that the ranking model's induced policy maximizes the online metrics upon deployment.

\subsection{The Lower Level: Model Training}
We consider a standard recommendation setting where a ranking model, parameterized by weights $\theta$, ranks a list of candidate items to generate an action $a$ (the ranking order) given state $s$, which comprises the user state and candidate videos, to maximize a cumulative reward. 

While the methodology proposed in this paper is model-agnostic, our deployment environment utilizes a value-based RL approach \cite{lrf_paper, deeprl_paper}. Specifically, the model optimizes a state-action value function $Q_\theta(s, a)$ that estimates the long-term value of a ranking action, defined by a proxy reward constructed from session-level user-item interactions. The model is trained via Stochastic Gradient Descent (SGD) to minimize a differentiable proxy loss function $\mathcal{L}_{\text{proxy}}$:

$$\theta^*(\Phi) = \arg\min_{\theta} \mathcal{L}_{\text{proxy}}(\mathcal{D}; \theta, \Phi)$$

where $\mathcal{D}$ represents the training data logs and $\Phi$ represents the meta-configuration of the system, such as the model optimizer, architecture, and reward definition. This lower-level optimization is performed by the model trainer.

\subsection{The Upper Level: Optimizing North Star Metrics}

While the ranking model optimizes the proxy reward, industrial recommendation systems ultimately care about the true north star metrics $\mathcal{M}$. The mapping between the proxy reward and the true online metrics is not guaranteed; a model might improve offline loss on a poorly defined reward function to the detriment of user satisfaction.

Thus, we formulate the problem of finding the optimal configuration $\Phi$ as:

$$\Phi^* = \arg\max_{\Phi}  \mathcal{M}(\theta^*(\Phi)) \quad \text{s.t.} \quad \mathcal{G}(\Phi) \le C$$

Here, $\theta^*(\Phi)$ represents the model weights trained under configuration $\Phi$, and $\mathcal{G}$ represents system-level constraints (e.g., training cost).

This formulation highlights the challenge: we must optimize $\Phi$ using expensive, noisy feedback from $\mathcal{M}$ to ensure the ranking model, which efficiently optimizes reward, is actually solving the business problem. Traditionally, optimizing $\Phi$ has been undertaken by human researchers. Our goal is to automate the role of a human researcher via an MLE agent that iteratively refines the components of $\Phi$. Concrete examples of $\Phi$ include:

\begin{itemize}
    \item \textbf{Optimizer ($\eta \in \Phi$)} The learning rate and update rules (e.g., AdaGrad) used to train the model weights $\theta$.
    \item \textbf{Architecture ($\phi \in \Phi$)} The structure (e.g., DCN) of the ranking network.
    \item \textbf{Reward Definition ($r \in \Phi$)} The logic determining the training labels, combining various engagement and user signals to balance competing objectives.
\end{itemize}

\section{The Self-Evolving System}
We propose a system that automates the discovery of an optimal model configuration $\Phi^*$ by decoupling the discovery process into two distinct, synchronized reasoning loops. The system is designed around the Experiment Journal -- a shared, persistent knowledge base containing the history of all configurations, their offline scores, and any available online metrics -- which informs two primary agents (Figure \ref{fig:system_architecture}):

\begin{figure*}[ht]
\centering
\includegraphics[width=0.85\textwidth]{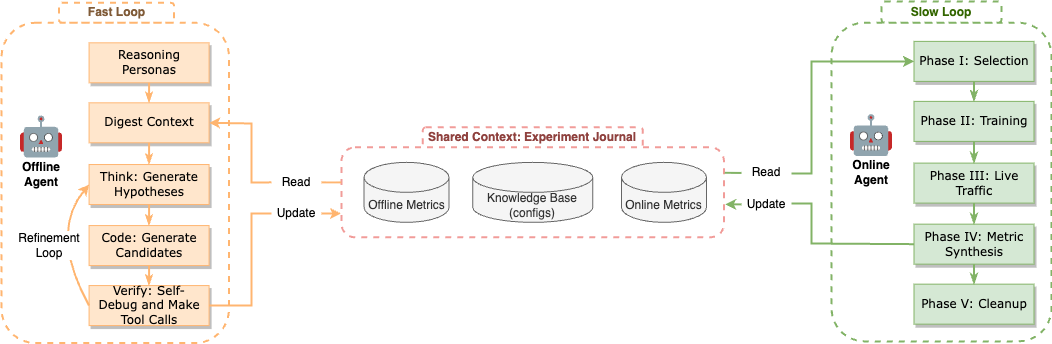}
\caption{The Self-Evolving System Architecture. The framework operates as a dual-loop, self-evolving system centered around a shared Experiment Journal containing a persistent knowledge base, historical trials and their resulting metrics. The Offline Agent (Fast Loop) serves as the high-frequency nomination engine, where LLMs are invoked to instantiate specialized reasoning personas that generate and refine hypotheses into executable code. Tool calls are made to assign offline scores to candidates. The Online Agent (Slow Loop) is the low-frequency ranking engine, selecting high-potential candidates and promoting them to online experiment. It manages the entire experiment lifecycle including fetching online north star metrics.}
\Description{A flowchart showing the architecture, including the Offline Agent loop and the Online Agent loop.}

\label{fig:system_architecture}
\end{figure*}

\begin{figure}[t]
\centering
\fbox{
\begin{minipage}{0.96\columnwidth}
\textbf{\# GOAL}\\
Propose changes to the model, with the following goals: \\
- Balance exploration, exploitation, and innovation: You should make X, Y, and Z proposals, respectively, in the three categories.\\
- Minimize the offline metric: \texttt{\{AGENT\_OFFLINE\_METRIC\}}, while maximizing the following online metrics in order of importance: Metric\#1,  Metric\#2, .. \\

\textbf{\# OUTPUT FORMAT} \\
Think step-by-step and double-check syntax. Output each proposal with exactly two fields: \\
- "explanation", briefly describing what change this is and why it is potentially useful \\
- "delta", the change against the model's current configuration

\end{minipage}
}
\captionof{figure}{Offline Agent Task. This text is injected into the shared prompt template in Figure \ref{fig:prompt_template} as the \texttt{\{AGENT\_TASK\}}.}
\label{fig:offline_goal}
\Description{Offline Agent Task.}
\end{figure}

\begin{figure}[t]
\centering
\fbox{
\begin{minipage}{0.96\columnwidth}
\textbf{\# GOAL}\\
Propose a rank of the top $K$ configurations in the history that are expected to maximize the online performance. \\
- The offline metric is \texttt{\{AGENT\_OFFLINE\_METRIC\}}, with a smaller value indicating better offline performance. \\
- The online metrics in order of importance are: Metric\#1,  Metric\#2, .. \\

\textbf{\# GUARDRAILS}\\
Maintain system safety by enforcing: Keep Metric\#3 $\leq +1\%$, .. \\

\textbf{\# OUTPUT FORMAT} \\
Output an ordered list of $K$ configurations, starting with the most promising configuration. Each item in the list must have exactly two fields: \\
- "name", an identifier for the configuration \\
- "delta", the change from the history
\end{minipage}
}
\captionof{figure}{Online Agent Task. This text is injected into the shared prompt template in Figure \ref{fig:prompt_template} as the \texttt{\{AGENT\_TASK\}}.}
\label{fig:online_goal}
\Description{Online Agent Task.}
\end{figure}

\begin{enumerate}
\item The Offline Agent (Fast Loop): Operates as a high-frequency candidate generator, running once every day but waking up every 5 minutes, tasked with \textbf{nominating} potential model improvements. Each wakeup performs one or more of: nominating new candidates, scheduling new training jobs or analysis queries based on the latest day's user data, and scoring candidates. Its goal is to identify configurations that minimize the offline metric (e.g., loss, inverted correlation) within regions of the configuration space that have historically yielded promising online metrics (see the agent-specific prompt in Figure \ref{fig:offline_goal}). The agent uses available live signals to steer away from dead-end trajectories that appear promising in offline proxies but have historically failed to improve online metrics.

\item The Online Agent (Slow Loop): Operates as a low-frequency experiment strategist, running once every day, tasked with \textbf{ranking} candidates produced by the Fast Loop. Its goal is to identify candidates that maximize the north star metrics $\mathcal{M}$ (see the agent-specific prompt in Figure \ref{fig:online_goal}). This agent uses the rank to decide which configurations merit promotion to live experiments and when to terminate unpromising experiments, ensuring that expensive online traffic is reserved for candidates with the highest demonstrated potential.

\end{enumerate}

Both agents utilize a shared prompt template, with the specifics of each agent injected into it (Figure \ref{fig:prompt_template}). The prompt structure integrates the agent's goals with the Experiment Journal, enabling each agent to reason over historical successes and failures. Additionally, we use few-shot prompting by including example proposals to guide the LLM's output format.

\begin{figure}[t]
\centering
\fbox{
\begin{minipage}{0.96\columnwidth}
\textbf{\# PERSONA}\\
You are a brilliant and innovative machine learning scientist with excellent programming and analytical skills. You want to improve the model and you have deep expertise in \texttt{\{AGENT\_SPECIALIZATION\}}. \\

\texttt{\{AGENT\_TASK\}} \\

\textbf{\# CONTEXT}\\
The model currently has the following configuration: \texttt{[BASELINE CONFIGURATION]} \\
(Optional) The output (hypotheses and data) from past data analyses are: \texttt{\{SQL\_QUERY\_OUTPUT\}} \\
Below is the history of past offline and (if available) online results, sorted by the offline score where the best offline configuration is shown first: \texttt{[EXPERIMENT JOURNAL]} \\

\textbf{\# EXAMPLE PROPOSAL} \\
\texttt{\{AGENT\_EXAMPLE\}}
\end{minipage}
}
\captionof{figure}{Shared LLM Prompt Template. \texttt{\{..\}} are populated based on the agent's task, and \texttt{[..]} are populated from shared context.}
\label{fig:prompt_template}
\Description{Shared LLM Prompt Template.}
\end{figure}

\subsection{The Offline Agent: A Fast Loop}
The Offline Agent serves as the rapid nomination engine of the framework, responsible for generating high-potential candidates for experimentation by the Online Agent. It utilizes specialized reasoning personas to traverse the semantic configuration space guided by historical metrics. Furthermore, the agent leverages a suite of tools to assign offline scores to the generated candidates.

\subsubsection{Specialized Reasoning Personas and Tool Calls}
We decompose the multifaceted task of recommendation system design into specialized agent personas, each optimized for a distinct model component. A monolithic persona exposed to the full breadth of the codebase quickly becomes overwhelmed by irrelevant schema definitions and knowledge, leading to hallucinations that we observed starting at \textasciitilde{400k} tokens. We thus instantiate  specialized personas, each equipped with specific tools and objectives and focusing on improving $\Theta(10)$ lines of a component's code within $\Theta(1000)$-line model definition -- typical of the size of a human-engineered change.

\textbf{A. The Optimizer Persona (Tool: \texttt{compute\_loss})} 
This persona searches the space of optimization algorithms. It iteratively proposes changes to the optimizer class (e.g., Adagrad, RMSprop) and its internal hyperparameters (e.g., momentum, batch size). 

For such changes, the definition of the loss function $\mathcal{L}_{\text{proxy}}$  remains invariant. While the model changes, the yardstick measuring its performance does not. Consequently, the resulting validation loss values are comparable. A lower loss implies a better approximation of the ground truth labels.

\begin{itemize}
    \item \textbf{Objective} Minimize offline loss $\mathcal{L}_{\text{proxy}}$.
    \item \textbf{Tooling} The persona utilizes the \texttt{compute\_loss} tool. This enables a direct sorting of candidates: $\Phi_A \succ \Phi_B \iff \mathcal{L}_{\text{proxy}}(\Phi_A) < \mathcal{L}_{\text{proxy}}(\Phi_B)$.
    \item \textbf{Process} The persona generates multiple configurations (e.g., replacing optimizers, changing learning rates) and launches asynchronous training jobs.
    \item \textbf{Scoring} The persona computes the loss for each configuration.
\end{itemize}

\textbf{B. The Architecture Persona (Tool: \texttt{compute\_loss})} This persona specializes in the neural topology of the model, parsing the architecture definition and proposing structural mutations. Unlike standard NAS, which selects from a fixed menu of layers, this persona can write novel code -- for example, replacing standard embedding lookups with a custom "Gated Path" mechanism or introducing layer normalization in specific sub-towers (discussed in Section~\ref{sec:loss_architecture_tasks}). Like the Optimizer persona, it relies on the \texttt{compute\_loss} tool to score the configuration, where lower scores indicate a topology that is more learnable and expressive than the baseline.

\textbf{C. The Reward Persona (Tool: \texttt{run\_sql\_query}, \texttt{compute\_eval})} 
This persona performs reward engineering by editing the logic that defines the training label for the ranking model. It utilizes a multi-step reasoning process to discover and validate new rewards.

First, it performs open-ended and massive-scale data analyses using the \texttt{run\_sql\_query} tool, finding hypotheses and relationships in the user logs. An example finding is the hypothesis that "videos that are shared yield higher watch time", backed by data comparing the average watch time of videos that are shared vs not shared.

Second, it generates configurations based on these findings (injected into the prompt via \texttt{\{SQL\_QUERY\_OUTPUT\}} in Figure \ref{fig:prompt_template}), and computes loss-independent proxy metrics for them. Because modifying the reward fundamentally alters the optimization landscape, comparing $\mathcal{L}_{\text{proxy}}$ across different reward definitions is ill-defined. A model trained on a "click-only" reward will naturally have a lower loss than one trained on a complex "click + satisfaction" reward, simply because the latter task is harder to learn. Therefore, this persona cannot use \texttt{compute\_loss}. Instead, it relies on the \texttt{compute\_eval} tool to calculate a proxy metric that is highly predictive of desirable user behavior using the evaluation data. An example is the long-watch correlation, which represents how well the model's prediction aligns with the actual occurrence of long watches, with a higher score indicating that the model is successfully prioritizing content that users find engaging enough to watch above a duration. More examples (e.g., correlation with retention and repeat consumption) can be found in \cite{surrogate_paper}. 

\begin{itemize}
\item \textbf{Objective} Identify promising signals that are highly predictive of user engagement.
\item \textbf{Tooling} The persona utilizes the \texttt{run\_sql\_query} tool to do open-ended data analyses, and the \texttt{compute\_eval} tool to validate the quality of the new reward.
\item \textbf{Process} The agent executes batches of analytical queries over user logs, and launches training jobs for the proposed configurations.
\item \textbf{Scoring} The persona scores each configuration by computing a surrogate proxy.
\end{itemize}

\subsection{The Online Agent: A Slow Loop}

The Online Agent prioritizes the accuracy and safety required for production deployment by selecting the top candidates produced by the Offline Agent and validating them against delayed north star metrics $\mathcal{M}$. It manages the experiment lifecycle of the top candidates through five phases, beginning with a reasoning phase to determine the trajectory of each candidate:

\begin{enumerate}
  \item \textbf{Selection}: The Online Agent begins by evaluating the entire candidate pool stored in the Experiment Journal. Its task is to rank the candidates to identify the top $K$ for online experimentation. Importantly, while the Experiment Journal is ordered by offline metrics, the Online Agent must also consider the online metrics to produce a refined ranking. The resulting LLM output categorizes the candidates into three distinct paths:
  \begin{itemize}
    \item Candidates newly in the top $K$ are routed through phases (2) and (3) to begin their training and experimentation.
    \item Candidates that are already in active experiments and remain in the top $K$ are routed to phase (4) for continued metric collection.
    \item Candidates that have fallen out of the top $K$ are routed to phase (5) for resource reclamation.
\end{itemize}
  \item \textbf{Model Training}: The agent trains new configurations and monitors convergence to ensure weights are successfully versioned and exported. 
  \item \textbf{Live Experimentation}: The agent assigns production traffic to the newly trained models to begin online experimentation.
  \item \textbf{Metric Synthesis}: The agent fetches online metrics and writes them into the Experiment Journal. This data is critical for the next iteration's reasoning.
  \item \textbf{Cleanup}: For candidates no longer in the top $K$, the agent reclaims resources by cleaning up their trainers and experiments. This closes the loop for unpromising directions.
\end{enumerate}

By employing a dual-agent system instead of a monolithic agent, we establish a rigorous filtration funnel that distinguishes high-velocity candidate nomination from strategic experimentation. In this new methodology, human engineers are only required to perform the high-level step of presenting the initial research idea to the Offline Agent and the final step of reviewing experiment metrics collected by the Online Agent.

\begin{table*}[ht]
\centering
\caption{Agent Components vs Discovered Improvements vs Online Metrics (for a primary YouTube surface)}
\label{tab:online_summary}
\begin{tabular}{@{}llcc@{}}
\toprule
\textbf{Component} & \textbf{Discovery} & \textbf{YouTube-level metric} & \textbf{Surface-level metric} \\ \midrule
\textbf{Optimizer} & Transition to RMSprop & \textbf{+0.06\% [+0.03\%, +0.09\%]} & \textbf{+0.12\% [+0.05\%, +0.19\%]} \\
\textbf{Optimizer} & Training Efficiency ($4\times$ Improvement) & $-0.01\% [-0.05\%, +0.03\%]$ & $+0.06\% [-0.02\%, +0.13\%]$ \\
\textbf{Optimizer} & Training Efficiency ($2\times$ Improvement) & $+0.01\% [-0.03\%, +0.05\%]$ & \textbf{+0.09\% [+0.04\%, +0.15\%]} \\ \midrule
\textbf{Architecture} & Gated Path (GLU) & \textbf{+0.06\% [+0.02\%, +0.11\%]} & \textbf{+0.14\% [+0.08\%, +0.21\%]} \\
\textbf{Architecture} & Activation Refinement & $-0.02\% [-0.05\%, +0.01\%]$ & \textbf{+0.12\% [+0.05\%, +0.19\%]} \\ \midrule
\textbf{Reward} & Multi-Objective Synthesis & \textbf{+0.05\% [+0.02\%, +0.08\%]} & \textbf{+0.17\% [+0.13\%, +0.22\%]} \\
\textbf{Reward} & Hyperparameter Tuning & \textbf{+0.05\% [+0.01\%, +0.08\%]} & \textbf{+0.21\% [+0.13\%, +0.29\%]} \\
\bottomrule
\end{tabular}
\par
\smallskip
{\footnotesize Results that are statistically significant at the 95\% confidence level are shown in bold.}
\end{table*}

\begin{table*}[ht]
\centering
\caption{Agent Components vs Discovered Improvements vs Online Metrics (for a different YouTube surface)}
\label{tab:online_summary_2}
\begin{tabular}{@{}llcc@{}}
\toprule
\textbf{Component} & \textbf{Discovery} & \textbf{YouTube-level metric} & \textbf{Surface-level metric} \\ \midrule
\textbf{Optimizer} & Transition to FTRL & \textbf{+0.03\% [+0.01\%, +0.05\%]} & \textbf{+0.16\% [+0.06\%, +0.26\%]} \\ \midrule
\textbf{Architecture} & Wide \& Deep Model & \textbf{+0.08\% [+0.04\%, +0.13\%]} & \textbf{+1.10\% [+0.98\%, +1.22\%]} \\
\bottomrule
\end{tabular}
\par
\smallskip
{\footnotesize Results that are statistically significant at the 95\% confidence level are shown in bold.}
\end{table*}

\section{Deployment and Results}

The self-evolving recommendation system has been deployed across several critical surfaces on YouTube. We present a comprehensive evaluation comparing our autonomous dual-agent system against human-engineered baselines.

\subsection{Experimental Setup}
We evaluate the efficacy of our framework in two stages, aligning with the dual-agent methodology we established. The first stage is offline validation via the Fast Loop, showing that LLM agents are capable of finding candidates that minimize loss or exhibit high correlation with key signals. The second stage is online A/B experimentation via the Slow Loop, showing that candidates reaching this stage significantly improve north star metrics. The underlying production model is an RL model based on a deep neural network to optimize video ranking on YouTube's video watch page, with training typically requiring $\Theta(hours)$. Final performance is judged against a hierarchy of north star business metrics $\mathcal{M}$.

A summary of the impact achieved by our autonomous agents is presented in Table~\ref{tab:online_summary}, with detailed discussion provided in Section~\ref{sec:loss_tasks} and Section~\ref{sec:reward_tasks}. To contextualize these gains, we analyzed all launches on our surface from the past 6 months. On average, the improvements generated by the agentic system outperformed \textbf{64\%} of launches generated using the traditional, manual approach for the YouTube-level metric and \textbf{73\%} for the surface-level metric.

\begin{table*}[t]
\centering
\small
\caption{Agentic Discovery Highlights: Evolution of Model Components}
\label{tab:evolution_highlights}
\begin{tabular}{llp{0.30\textwidth}p{0.34\textwidth}}
\toprule
\textbf{Component} & \textbf{Discovery} & \textbf{Initial Configuration} & \textbf{Evolved Configuration} \\
\midrule
\textbf{Optimizer} & Transition to RMSprop & \texttt{Adagrad(learning\_rate=0.1)} & \texttt{RMSprop(learning\_rate=0.005,rho=0.95,...)} \\
\midrule
\textbf{Architecture} & Gated Path (GLU) & \texttt{layer\_norm(relu(dense(relu(dense(
inputs,128,'linear'),128,'linear')))} & \texttt{value\_path*gate\_path, where:} \newline
\texttt{value\_path=dense(shared,32,'linear'),} \newline \texttt{gate\_path=dense(shared,32,'sigmoid'),} \newline  \texttt{shared=relu(dense(inputs,128,'linear'))} \\
\midrule
\textbf{Reward} & Multi-Objective Synthesis & \texttt{(factorA+factorB)*exprC} & \texttt{factor\_new*(factorA+factorB)*exprC, where:} \newline
\texttt{factor\_new=GREATEST(...,1.0-GREATEST(0.0, CAST(IFNULL(...,0) AS FLOAT)-...)/...)}\\
\bottomrule
\end{tabular}
\end{table*}

\subsection{Evaluation of the Optimizer and Architecture Components: Loss Optimization}
\label{sec:loss_tasks}
The first phase of deployment focused on improving the optimizer and architecture, which both seek to minimize offline loss $\mathcal{L}_{\text{proxy}}$. The Offline Agent proposed the below refinements, which yielded significant improvements when promoted to the live environment.

\subsubsection{Algorithmic Discovery: Evolving the Optimizer}
Traditionally, optimizer configurations remained static due to the cost of tuning. The agent autonomously identified that switching from the legacy Adagrad optimizer to RMSprop (see Optimizer component in Table~\ref{tab:evolution_highlights}) -- with a specific learning rate, decay rate, momentum, etc. -- resulted in a statistically significant drop in offline loss and improvement in live traffic. Notably, the benefit of an LLM is we can simply ask it to find the "best Keras optimizer" without specifying what's available, as keywords are not the same for each optimizer.

\subsubsection{System Optimization: Training Efficiency}
Beyond model quality, the agent also learned to optimize for system efficiency. By iteratively adjusting batch sizes, training epochs, and optimizer hyperparameters, the agent achieved reductions -- first by $4\times$ then by $2\times$ -- in training latency without degrading convergence. In total, training time improved by $8\times$ without sacrificing business metrics.

\subsubsection{Structural Discovery: Gated Path Architectures and Activation Refinement}
\label{sec:loss_architecture_tasks}
After exploring hundreds of potential solutions to optimize the topology, the agent proposed a Gated Path architecture (see Architecture component in Table~\ref{tab:evolution_highlights}) similar to Gated Linear Units (GLU) \cite{glu_paper}, which introduced a multiplicative gate to the inputs. This innovation yielded some of the most robust gains in our deployment. In a follow-up deployment, the agent further refined this architecture by moving from standard sigmoid gates to GELU activations combined with layer normalization, showing that the agent is capable of both exploring innovative structures and exploiting and fine-tuning structures that it believes to be superior.

\subsection{Evaluation of the Reward Component: Semantic Alignment via Signal Correlation}
\label{sec:reward_tasks}
Unlike the Optimizer and Architecture improvements, which minimize the loss, the Reward improvement must interpret a loss-independent proxy to find a reward that balances conflicting business objectives while capturing nuanced user behaviors.

\subsubsection{Semantic Discovery: Multi-Objective Reward Synthesis}
Leveraging iterative analysis of user interaction patterns, the agent synthesized a reward function that incorporates a novel signal indicating whether the user is actively engaging with content on the site (see Reward component in Table~\ref{tab:evolution_highlights}). This synthesis significantly outperformed the human-engineered baseline -- a remarkable feat given the historical difficulty of manual reward engineering. There are $\Theta(100)$ signals and combinations -- watch time, survey responses, retention metrics, and more -- to consider to approximate long-term user satisfaction. Human researchers often struggle to pinpoint critical semantic bottlenecks within this massive search space, frequently resulting in months of iteration in suboptimal regions of the potential solution space. This discovery underscores the agent's unique ability to perform high-level semantic reasoning, allowing it to redefine the business logic of success in ways that traditional optimization processes cannot achieve.

\subsubsection{Reward Hyperparameter Tuning}

Beyond discovering novel reward structures, the agent demonstrated the ability to tune existing reward hyperparameters. This tuning was performed without offline metrics, relying exclusively on the Online Agent's slow loop to explore the parameter space. 
This is significant because previous manual attempts over several months to tune these hyperparameters failed to find a configuration that simultaneously improved both YouTube-level and surface-level metrics. The agent, however, identified a solution tuning four hyperparameters in two weeks.

\subsection{LLM Performance and Ablation Studies}
To understand the drivers of the system's performance, we conducted a series of ablation studies focusing on model selection, persona framing, and context management. These benchmarks highlight the sensitivity of the discovery process to the underlying LLM's reasoning capabilities and the quality of prompt grounding. We consider the task of improving the offline loss of the optimizer and its hyperparameters, using the following variants:
\begin{enumerate}
    \item \textbf{opt\_2p5 (Baseline)} Uses Gemini 2.5 Pro with an expert MLE persona framing, and the full history of past configurations and metrics sorted by offline loss.
    \item \textbf{opt\_flash} Uses Gemini 2.5 Flash instead of Pro.
    \item \textbf{opt\_no\_role} Ablates the expert MLE persona framing.
    \item \textbf{opt\_no\_sort} Provides the full history of past metrics but ordered by timestamp instead of loss.
    \item \textbf{opt\_top\_1 / opt\_top\_5} Limits the history to only the top 1 and 5, respectively, sorted by offline loss.
    \item \textbf{opt\_no\_context} Provides no history of past configurations or metrics.
\end{enumerate}

Results are averaged over 6 independent runs exploring 70 ideas each, reported as normalized z-scores of the loss where lower (more negative) values indicate superior performance (Table~\ref{tab:agent_ablation}).

\begin{table}[ht]
\centering
\caption{Agent Performance (Normalized $\mathcal{L}_{\text{proxy}}$) for the Optimizer Component}
\label{tab:agent_ablation}
\begin{tabular}{lc}
\toprule
\textbf{Agent Configuration} & \textbf{Normalized Loss} \\
\midrule
opt\_2p5 & $-0.84$ $[-1.70, -0.01]$ \\
opt\_top\_5 & $-0.72$ $[-1.49, 0.05]$ \\
opt\_no\_role & $-0.52$ $[-1.43, 0.38]$ \\
opt\_no\_sort & $0.06$ $[-0.93, 1.05]$ \\
opt\_top\_1 & $0.11$ $[-0.76, 0.98]$ \\
opt\_flash & $0.85$ $[0.66, 1.05]$ \\
opt\_no\_context & $1.05$ $[0.94, 1.16]$ \\
\bottomrule
\end{tabular}
\end{table}

\subsubsection{Impact of Model Size and Reasoning}
\label{sec:gemini_benchmark}
We evaluated the effectiveness of different model choices within the Gemini family. As shown in Table \ref{tab:agent_ablation}, a larger model with advanced reasoning capabilities significantly outperforms a smaller variant. Specifically, Gemini 2.5 Pro consistently achieves lower loss compared to Gemini 2.5 Flash. This confirms that the reasoning required for algorithmic discovery benefits from the increased parameter count and enhanced "deep thinking" capabilities of the larger model class.

\subsubsection{Persona and Context Length}
We evaluated the impact of the expert MLE persona framing by comparing it against an agent that lacked the expert identity. This comparison confirms that expert framing influences the relevance and depth of proposed configurations, making it critical for model quality. Similarly, context engineering plays a vital role: providing the full, sorted history from the Experiment Journal is better than no history, restricted top-k, or unsorted history. As seen in Table \ref{tab:agent_ablation}, these results suggest that a comprehensive and ranked distribution of past outcomes is essential for effective iterative discovery.

\subsection{Efficiency and Costs}
A critical consideration for autonomous discovery systems is the trade-off between the velocity gains enabled by the system and the associated operational costs.

\subsubsection{The Velocity Dividend}
By decoupling the high-frequency offline discovery from the low-frequency online validation, we have removed human engineers from the repetitive and manual path. The "Idea-to-Data" cycle -- the latency from hypothesis generation to experimental results -- is significantly cheaper and faster. This velocity dividend now enables the team to produce far more launches than previously possible (Table~\ref{tab:efficiency}).

\subsubsection{LLM \& Infrastructural Costs}
The LLM token costs were approximately \$20,000 over a six-month period, representing a minimal fraction of the cost of a full-time MLE. The per-model compute and memory costs remain identical for both human-driven and agentic workflows, so the remaining infrastructure costs scale directly with the number of models tested.

\begin{table}
\centering
\caption{Experimental Velocity vs Type of Workflow}
\label{tab:efficiency}
\begin{tabular}{l|cc}
\toprule
\textbf{Metric} & \textbf{Human Workflow} & \textbf{Agent Workflow} \\
\midrule
Exp. Throughput & $\Theta(1) - \Theta(10)$ / week & $\Theta(100)$ / week \\
Eng. Cost per Exp & $\Theta(1) - \Theta(10)$ hours / week & $0$ hours / week \\
\bottomrule
\end{tabular}
\end{table}

\subsection{Lessons Learned}
The deployment of an autonomous, self-evolving system provided several critical insights into the future of recommendation system engineering, ranging from the practical considerations essential for production stability (\textbf{L1}-\textbf{L3}) to the transformative reasoning power unlocked by an agentic system (\textbf{L4}-\textbf{L5}).

\begin{itemize}
    \item \textbf{L1: Delta-based vs. Full Configuration Generation} The validity of proposals was significantly enhanced when the agent was tasked with generating a delta against the production file rather than the entire configuration file. Requesting the complete configuration often led to hallucinations where the model would omit essential but unchanged parameters or introduce syntax errors. Furthermore, we find delta-based generation to offer a decisive advantage in context management. Because deltas are compact, we can attach $\Theta(100)$ items from past history as context to each prompt. In contrast, existing methods like AlphaEvolve must aggressively trim context because it is not feasible to include that much history when the complete files are large.
    \item \textbf{L2: Enforcing Diversity via Prompt Tuning} Absent explicit instructions, the agent exhibited a strong bias toward safe, incremental changes, effectively collapsing into a mode of minor hyperparameter tuning (e.g., proposing "learning rate 0.1" followed immediately by "learning rate 0.11"). To counteract this, it was critical to prompt the agent to "balance exploration, exploitation, and innovation", forcing it to attempt the leaps necessary for significant gains.
    \item \textbf{L3: Importance of Warm Start} The ability to nominate good configurations is highly dependent on the density of the Experiment Journal. Without past trials, the agent tends to propose generic textbook improvements. To jumpstart the discovery, we warm-start the Offline Agent's daily run with the best $K$ findings from the preceding day, allowing the agent to ground its reasoning in past learnings.
    \item \textbf{L4: Semantic Reasoning vs. Numerical Tuning} While traditional AutoML excels at tuning scalars, our results show that the highest leverage in mature systems comes from structural and semantic mutations. For example, the Reward persona's ability to redefine the reward provided innovations that purely numerical tuning could never achieve.
    \item \textbf{L5: Generalizability Across Recommendation Surfaces} A critical question for any autonomous framework is transferability: can the agent adapt to new environments without re-engineering? We deployed the same dual-agent architecture to a different YouTube recommendation surface with a completely different feature schema, training dataset, and model configuration. Despite these differences, the agents successfully adapted to the new context within the first few iterations, generated valid hypotheses, and increased north star metrics (Table~\ref{tab:online_summary_2}). This confirms that our framework optimizes the process of discovery rather than memorizing a specific dataset, suggesting strong potential for generalization across the broader family of recommendation systems.
\end{itemize}

\section{Conclusion}
In this paper, we present a comprehensive framework leveraging Large Language Models (LLMs) for a self-evolving recommendation system, successfully deploying it at scale on the world's largest video delivery platform. By decoupling the discovery process into a fast Offline Agent (driven by cheap proxy signals) and a reliable but slow Online Agent (driven by delayed north star business metrics), we have established a new paradigm for industrial machine learning that overcomes the limitations of traditional workflows. Our extensive deployment results illustrate critical contributions of our work to the field of automated machine learning. We showed that LLMs, when grounded with the appropriate context and tools, are capable of structural and semantic innovation in recommendation systems. And by automating the repetitive mechanics of code generation, compilation, and experiment orchestration, we noticeably compressed the "Idea-to-Data" cycle. This order-of-magnitude increase in experimental throughput allows the system to explore the long tail of the configuration space that human engineers simply do not have the bandwidth to investigate. 

Looking forward, we envision a shift in the role of the Machine Learning Engineer (MLE). As a self-evolving recommendation system executes on modeling improvements, the human engineer moves focus to defining the strategic guardrails, ethical constraints, and the long-term vision of the system. We believe our work represents a foundational step toward that future, removing human cognitive bandwidth as a bottleneck in scientific discovery in recommendation systems.

\bibliographystyle{ACM-Reference-Format}
\bibliography{references}

\end{document}